\documentclass[sigconf]{acmart}

\AtBeginDocument{%
  }

\copyrightyear{2026}
\acmYear{2026}
\setcopyright{cc}
\setcctype{by}
\acmConference[KDD '26]{Proceedings of the 32nd ACM SIGKDD Conference on Knowledge Discovery and Data Mining V.2}{August 09--13, 2026}{Jeju Island, Republic of Korea}
\acmBooktitle{Proceedings of the 32nd ACM SIGKDD Conference on Knowledge Discovery and Data Mining V.2 (KDD '26), August 09--13, 2026, Jeju Island, Republic of Korea}
\acmDOI{10.1145/3770855.3816449}
\acmISBN{979-8-4007-2259-2/2026/08}

\usepackage{latexsym}

\usepackage{enumitem}
\usepackage{balance}

\usepackage[T1]{fontenc}

\usepackage[utf8]{inputenc}

\usepackage{microtype}

\usepackage{graphicx}

\title{Agents in the Wild: Where Research Meets Deployment}

\author{Grace Hui Yang}
\email{grace.yang@georgetown.edu}
\affiliation{%
  \institution{Georgetown University}
  \city{Washington}
  \state{D.C.}
  \country{USA}
}
\authornote{Authors are listed in inverse alphabetical order.}
\author{Pranav N. Venkit}
\email{pnarayananvenkit@salesforce.com}
\affiliation{%
\institution{Salesforce}
  \city{San Francisco}
  \state{California}
  \country{USA}
}
\author{Hooman Sedghamiz}
\email{hooman.sedghamiz@bayer.com}
\affiliation{%
  \institution{Bayer}
  \city{San Diego}
  \state{California}
  \country{USA}
}
\author{Enrico Santus}
\email{esantus@bloomberg.net}
\affiliation{%
  \institution{Bloomberg}
  \city{New York City}
  \state{New York}
  \country{USA}
}
\author{Victor Dibia}
\email{victordibia@microsoft.com}
\affiliation{%
  \institution{Microsoft Research}
  \city{Redmond}
  \state{Washington}
  \country{USA}
}
\author{Ioana Baldini}
\email{ibaldinisoar@bloomberg.net}
\affiliation{%
  \institution{Bloomberg}
  \city{New York City}
  \state{New York}
  \country{USA}
}

\renewcommand{\shortauthors}{Grace Hui Yang et al.}

\begin{document}
\begin{abstract}
Agentic systems — large language model (LLM)-based architectures capable of reasoning, planning, acting, and coordinating with tools and other agents — are rapidly transitioning from research prototypes to production-scale deployments across domains such as software engineering, scientific discovery, and finance.
While academic work has emphasized benchmarks and algorithmic innovation, deployment raises new challenges around robustness, safety, and reliability. This tutorial brings together researchers and practitioners to explore advances in reasoning and planning, multi-agent coordination, and evaluation, highlighting open challenges arising from deployment experience. 
Through applied case studies in pharmaceutical discovery and financial systems, we analyze common design patterns that make agentic systems successful, and discuss practical mitigation strategies for failure modes, such as verification pipelines, fallback mechanisms, and human-in-the-loop supervision.
Attendees will gain a comprehensive view of the field along with concrete design patterns, evaluation checklists, and templates for safe and reliable deployment across industries.
\end{abstract}



\maketitle

\section{Tutorial Description}

In early work, agents were built as single-model pipelines combining prompting, tool use, and heuristic control~\cite{yao2023react}. As tasks and interaction environments grew more complex, the field has shifted toward modular, multi-agent architectures~\cite{du-etal-2025-multi, mozannar2025magentic, multiagentbench2025}. This tutorial traces the evolution from monolithic prompting-based agents to modular, multi-agent systems, illustrating how reasoning, planning, execution, and memory components interlock~\cite{zhang-etal-2025-planning}. 

To bridge research innovation and real-world deployment, we move beyond static evaluation metrics to examine how agents can be benchmarked in interactive environments. Deployment-focused case studies (e.g., autonomous data-analysis agents in finance, biomedical literature agents, and multi-agent code generation for software engineering) will ground the discussion with concrete analyses of failure modes — hallucination, deadlocks, drift, and cascading errors — and mitigation strategies such as cross-checking and human-in-the-loop fallback mechanisms. 

Finally, we highlight open research challenges, including multi-agent scalability, lifelong adaptation, explainability of collective reasoning, secure inter-agent communication, and embodied or multimodal agents, while emphasizing industrial constraints — latency, compute efficiency, and monitoring — and the need for continued academic rigor in designing safe, reliable agentic systems.

\section{Expected Audience Takeaways}
Participants will leave the tutorial with:

\begin{itemize}
  \item A clear understanding of the evolution of agentic systems — from single-LLM prompting to modular and multi-agent orchestration.
  \item An introduction to modern reasoning, planning, and execution strategies, including their trade-offs and representative examples.
  \item Evaluation techniques beyond static benchmarks, emphasizing robustness, reliability, safety, and continuous adaptation.
  \item Concrete case studies from deployment domains (e.g., bioinformatics pipelines, financial assistants) illustrating common failure modes and effective recovery strategies.
  \item Awareness of open research challenges and opportunities at the intersection of academic innovation and industrial practice.
\end{itemize}

\section{Tutorial Part I: Foundations}
\subsection{Agentic Systems: History and Definitions}

LLM research has rapidly evolved beyond simple generation, giving rise to sophisticated \textit{agentic systems} — autonomous, goal-directed entities that can reason, act, and adapt~\citep{zhao2025llmbasedagenticreasoningframeworks}. This transformation builds on the LLM’s reasoning capabilities~\citep{wei2022chainofthought, wang2023selfconsistency, yao2023treeofthoughts} by integrating external tools and interfaces~\citep{schick2023toolformer, Qu_2025}, enabling agents to perceive and interact with environments~\citep{browsergym2025} and to continually learn or update their knowledge~\citep{zheng2025towardslifelong,ning2025surveywebagents}. Current research and industry efforts are converging on three central challenges: (1) making individual agents more capable through enhanced reasoning and planning~\cite{wei-etal-2025-plangenllms}; (2) improving collective performance through multi-agent coordination~\citep{dang2025multiagent, tian2025strongestllmmultiturnmultiagent, tran2025multiagentsurvey}; and (3) ensuring robustness and safety during deployment~\citep{chaos2025, Mohammadi2025, huang2024understandingplanning}.

\subsection{Agent Foundations: Reasoning, Planning, and Multi-Agent Coordination}

To advance autonomous decision-making, researchers have developed techniques for complex reasoning and planning. Examples include task decomposition for breaking down goals~\citep{wang2025tdag, gabriel2024advancingagenticsystemsdynamic, liu2025toolplanner}, multi-plan generation and selection~\citep{parmar2025plangen}, iterative reflection loops that allow self-critique and revision~\citep{yuan2025agentr}, and memory-augmented planning for long-horizon tasks~\citep{shinn2023reflexion}. In parallel, the field is exploring how multiple LLM-based agents can collaborate or compete effectively, giving rise to \textit{multi-agent} orchestration strategies~\citep{dang2025multiagent, tian2025strongestllmmultiturnmultiagent, tran2025multiagentsurvey, fourney2024magentic}. These efforts investigate architectural topologies (peer-to-peer, hierarchical), communication protocols, negotiation versus cooperation schemes, and fault-tolerance mechanisms to coordinate teams of agents.

\subsection{Retrieval and Reasoning Pipelines}

As tasks become increasingly complex and interactive, simple one-shot knowledge retrieval proves insufficient for multi-step reasoning and dynamic adaptation. To address this, emerging agent pipelines employ \textit{iterative} retrieval strategies that interleave reasoning and information lookup~\citep{trivedi2023interleavingretrievalchainofthoughtreasoning}, \textit{adaptive} retrieval that learns when to consult external resources~\citep{asai2023selfraglearningretrievegenerate}, and \textit{modular} retrieval architectures that decompose the knowledge-search process into submodules (query generation, reranking, memory retrieval)~\citep{gao2023retrievalaugmented}. Such flexible retrieval-augmented reasoning pipelines enable agents to tailor knowledge access dynamically during decision-making, which is particularly crucial for interactive, open-ended tasks.

\subsection{Evaluation Beyond Benchmarks}

There are several benchmarks that have been developed to assess progress in agentic systems~\citep{webarena2024, swebench2024, drouin2024workarena, boisvert2024workarenaplusplus,huang2025crmarena,huang2025crmarenapro, browsergym2025,Mohammadi2025}. Recent work motivates a paradigm shift toward dynamic, behavior-centric evaluation frameworks that capture reliability and safety under real-world conditions~\cite{yehudai2025survey}. New efforts such as \textit{AgentSafetyBench}~\cite{zhang2025agentsafetybenchevaluatingsafetyllm} and \textit{ST-WebAgentBench} \cite{levy2024st} evaluate agent adaptability to distribution shifts, adversarial perturbations, and cascading failures, emphasizing recovery and human-intervention behaviors. Building on this, \citet{venkit2024searchenginesaiera} and \citet{venkit2025deeptrace} foreground the sociotechnical dimensions of agent assessment, with \textit{DeepTrace} \cite{venkit2025deeptrace} and \textit{Agent Security Bench (ASB)} \cite{zhangagent} formalizing societal impact and security resilience. Domain-specific extensions such as \textit{MobileSafetyBench} \cite{lee2024mobilesafetybench}, \textit{ScienceAgentBench} \cite{chen2025scienceagentbenchrigorousassessmentlanguage}, and \textit{AGENTSAFE} \cite{ying2025agentsafebenchmarkingsafetyembodied} expand safety evaluations across embodied and scientific contexts. These developments mark a transition from static evaluation to scenario-driven robustness, emphasizing behavioral safety and trustworthiness, which will be the key focus in this section of the tutorial.

\section{Tutorial Part II: Applied Perspectives}

\subsection{Agents in Pharmaceutical and Life Sciences}

The pharmaceutical and life sciences domain has become a leading testbed for agentic AI, transitioning from predictive modeling to fully autonomous scientific discovery spanning literature synthesis, hypothesis generation, and experimental validation. Recent breakthroughs such as FutureHouse’s ether0 \citep{narayanan2025ether0} and Robin \citep{ghareeb2025robin} showcase end-to-end discovery systems capable of superhuman molecular design and rapid drug candidate identification, while large-scale deployments like Moderna’s partnership with OpenAI report over 750 agentic applications across R\&D operations \citep{Lakhan2025TheAE}. These systems integrate specialized molecular modeling, reaction prediction, and lab automation capabilities \citep{boiko2023coscientist,bran2024chemcrow}, leveraging multi-agent architectures for literature analysis and experiment planning \citep{ghareeb2025robin,swanson2024virtuallab}, matching or exceeding human performance across complex discovery workflows \citep{narayanan2025ether0,ghafarollahi2024sciagentsautomatingscientificdiscovery}.


Multi-agent frameworks including PharmAgents \citep{gao2025pharmagents} and SciAgents \citep{ghafarollahi2024sciagentsautomatingscientificdiscovery} demonstrate collaborative discovery pipelines with measurable gains in success rate and material design innovation. Systems like Coscientist \citep{boiko2023coscientist}, ChemCrow \citep{bran2024chemcrow}, and Virtual Lab \citep{swanson2024virtuallab} validate outputs experimentally, while Google’s AI co-scientist \citep{gottweis2025aicoscientist} applies evolutionary debate mechanisms to uncover drug repurposing and epigenetic targets. Current work signals a new era of rapid, agent-driven biomedical discovery~\citep{gao2024empoweringbiomedicaldiscoveryai}.
Looking ahead, perspectives from Harvard Medical School \citep{gao2024empoweringbiomedicaldiscoveryai} envision AI scientists as collaborative, skeptical reasoners augmenting human creativity—heralding a new era of accelerated, agent-driven biomedical discovery. Biomedical researchers envision AI scientists as collaborative and critically reasoning partners that enhance human creativity, signaling a new era of accelerated, agent-driven biomedical discovery~\citep{gao2024empoweringbiomedicaldiscoveryai}.

\subsection{Agents in Finance}

In the financial domain, LLM-based agents are evolving beyond static analytical tools into dynamic decision-support systems and autonomous trading or analysis workflows. These systems commonly adopt a planner–executor–verifier architecture inspired by broader multi-agent orchestration frameworks such as AutoGen~\citep{wu2023autogenenablingnextgenllm,dibia-etal-2024-autogen}.
These agents combine natural language reasoning with real-time data retrieval and execution capabilities. For example, systems like \textit{FinAgent}~\citep{zhang2024multimodalfoundationagentfinancial} and \textit{FinRobot}~\citep{zhou2024finrobotaiagentequity} use multi-component architectures (LLM reasoning modules, financial data parsers, tool-using executors) to tackle tasks such as earnings call summarization, portfolio allocation, and risk assessment. By decomposing complex financial problems into sub-tasks and utilizing external data (market feeds, knowledge bases, calculators), they outperform single-step prompting methods on both decision quality and outcome explainability~\citep{yu2023finmemperformanceenhancedllmtrading}. 


\section{Tutors Bios}

{\bf Dr. Grace Hui Yang} is Professor of Computer Science at Georgetown University, where she leads the InfoSense (Information Retrieval and Sense-Making) research group. Her research bridges information retrieval and natural language processing, with a focus on conversational AI agents, retrieval-augmented generation, and deep reinforcement learning for dialogue and reasoning. Dr. Yang is a recipient of the NSF CAREER Award. She has served on the organizing and program committees of conferences (ACL, SIGIR, AAAI, CIKM, WSDM, and WWW), and on the editorial boards of ACM Transactions on Information Systems (2022–present) and the Information Retrieval Journal (2014–2017). She was General Co-Chair of SIGIR 2024 and has presented tutorials at SIGIR'14, WSDM'15, ICTIR'17, WSDM'18, WSDM'23, and SIGIR'23.

\medspace

\noindent{\bf Dr. Pranav N. Venkit} is a Research Scientist at Salesforce AI Research, where he contributes to the Interactive AI Team, building trustworthy, interactive agents capable of long-term reasoning, memory, and human–AI collaboration. His work currently focuses on understanding the impact of Agents in society and on how to build effective benchmarks to measure their use as sociotechnical systems. He earned his Ph.D. in Informatics from Penn State, where his dissertation, awarded Best AI Dissertation, explored sociotechnical language models through lenses of trust, fairness, and bias mitigation. His interdisciplinary research spans NLP, HCI, social informatics, and privacy, and has been published in venues such as ACL, EMNLP, EACL, and FAccT. 

\medspace

\noindent{\bf Hooman Sedghamiz} is Sr. Director of AI/ML at Bayer AG, where he leads the development of enterprise-scale generative AI and agentic systems for precision medicine serving more than 100,000 employees globally. 
He served as Co-Chair of the EMNLP 2023 GEM Industrial Track, establishing the inaugural industrial forum for the workshop. 
Prior to Bayer, he held research positions at University of California San Diego (AI/ML Advisor), LivaNova (Senior Research Scientist), and developed algorithms deployed in FDA-approved medical devices including cochlear implants, vagus nerve stimulators, and pacemakers. 
His work has been published in EMNLP, BMC Systems Biology, and Frontiers journals. He holds a granted international patent for medical device innovations and authored \href{https://joss.theoj.org/papers/10.21105/joss.00671}{BioSigKit}, an open-source biomedical signal processing toolkit with 45,240+ downloads. 
Hooman holds an M.Sc. in Biomedical Engineering from Linköping University, Sweden. 

\medspace

\noindent{\bf Dr. Enrico Santus} is a Principal Technical Strategist for QUANT NLP and Academic Engagement in Bloomberg's Office of the CTO. He has over 17 years of experience in developing and leading AI products across diverse domains including technology, medicine, pharmaceuticals, fake news detection, and finance. He holds a PhD in Computational Linguistics from The Hong Kong Polytechnic University, and has completed postdoctoral positions at SUTD and MIT. Enrico has authored over 70 conference and journal papers and organized numerous workshops (e.g., CMCL, GEM). In 2019 he was invited to speak about NLP at the White House, and he co-authored two editions of AI and ML factsheets for the US Congress. 

\medspace

\noindent{\bf Dr. Victor Dibia} is a Principal Research Software Engineer at Microsoft Research / Core AI where his contributions include \href{https://arxiv.org/abs/2507.22358}{Magentic-UI} (a framework for human-in-the-loop agentic systems), \href{https://arxiv.org/abs/2411.04468}{Magentic-One} (a generalist multi-agent system for solving complex tasks), \href{https://aclanthology.org/2024.emnlp-demo.8.pdf}{AutoGen Studio} (a no-code interface for prototyping multi-agent applications), \href{https://aclanthology.org/2023.acl-demo.11.pdf}{LIDA} (a system for automated data visualization, ACL 2023), and metrics for evaluating candidate Copilot models in production (ACL 2023). Prior to MSR, he was a Principal Research Engineer at Cloudera (Applied Machine Learning, Product) and a Research Staff Member at IBM Research New York (HCI and Applied ML Research). His work has been published in ACL, EMNLP, AAAI, and CHI, earning multiple best paper awards. 
Dr. Dibia holds a PhD in Information Systems from City University of Hong Kong and a Master's degree in Information Networking from Carnegie Mellon University. 

\medspace

\noindent{\bf Dr. Ioana Baldini} is a Research Scientist on the AI Strategy \& Research team within Bloomberg’s Office of the CTO, where she leads cross-functional collaborations and explores emerging technologies to advance AI innovation across the organization. Before joining Bloomberg, Ioana had a fruitful career at IBM Research that spanned multiple levels of the computational stack, including applied NLP, cloud infrastructure, and runtime systems for heterogeneous architectures. Ioana earned her master’s and Ph.D. from the University of Toronto. She received the NSERC Canada Graduate Scholarship, the IBM Ph.D. Fellowship, the Canada Google Anita Borg Scholarship, and the IBM Research Division Award. Ioana has presented tutorials in KDD'24, AAAI'25, NAACL'25, and FAccT'25. 

\section*{Acknowledgments}
This research was supported by U.S. National Science Foundation grant number IIS-2336768.

\balance

\bibliographystyle{ACM-Reference-Format}
\bibliography{wild_agents}




\end{document}